\documentclass[11pt]{article} \usepackage{acl2002}
\usepackage{times}
\usepackage{latexsym}
\title{Unsupervised discovery of morphologically related words based
on orthographic and semantic similarity}

\author{Marco Baroni\\
  \"OFAI\\
  Schottengasse 3\\
  A-1010 Vienna, Austria\\
  {\tt marco@oefai.at} \And
  Johannes Matiasek\\
  \"OFAI\\
  Schottengasse 3\\
  A-1010 Vienna, Austria\\
  {\tt john@oefai.at} \And
  Harald Trost\\
  IMKAI\\
  Freyung 6\\
  A-1010 Vienna, Austria\\
  {\tt harald@ai.univie.ac.at}}

\date{}

\begin{document}
\maketitle
\begin{abstract}
  We present an algorithm that takes an unannotated corpus as its
  input, and returns a ranked list of probable morphologically related
  pairs as its output. The algorithm tries to discover morphologically
  related pairs by looking for pairs that are both orthographically
  and semantically similar, where orthographic similarity is measured in
  terms of minimum edit distance, and semantic similarity is measured
  in terms of mutual information. The procedure does not rely on a
  morpheme concatenation model, nor on distributional properties of
  word substrings (such as affix frequency). Experiments with German and
  English input give encouraging results, both in terms of precision
  (proportion of good pairs found at various cutoff points of the
  ranked list), and in terms of a qualitative analysis of the types of
  morphological patterns discovered by the algorithm.
\end{abstract}

\section{Introduction}

In recent years, there has been much interest in computational models
that learn aspects of the morphology of a natural language from raw or
structured data.  Such models are of great practical interest as
tools for descriptive linguistic analysis and for minimizing the expert
resources needed to develop morphological analyzers and stemmers.  From a
theoretical point of view, morphological learning algorithms can
help answer questions related to human language acquisition.

In this study, we present a system that, given a corpus of 
raw text from a language, returns a ranked list of probable morphologically
related word pairs.  For example, when run with the Brown corpus as
its input, our system returned a list with pairs such as \emph{pencil/pencils}
and \emph{structured/unstructured} at the top.

Our algorithm is completely knowledge-free, in the sense that it 
processes raw corpus data, and it does not require any form of \emph{a priori} 
information about the language it is applied to. The algorithm performs 
unsupervised learning, in the sense that it does not require a 
correctly-coded standard to (iteratively) compare its output against.

The algorithm is based on the simple idea that a combination of 
formal and semantic cues should be exploited to identify 
morphologically related pairs. In particular, we use minimum edit 
distance to measure orthographic similarity,\footnotemark{} and mutual 
information to measure semantic similarity.
\footnotetext{Given phonetically transcribed input, our model would
  compute phonetic similarity instead of orthographic similarity.} The
algorithm does not rely on the notion of affix, and it does not depend
on global distributional properties of substrings (such as affix
frequency). Thus, at least in principle, the algorithm is well-suited
to discover pairs that are related by rare and/or non-concatenative
morphological processes.

The algorithm returns a list of related pairs, but it does not attempt
to extract the patterns that relate the pairs. As such, it can be
used as a tool to pre-process corpus data for an analysis to be performed
by a human morphologist, or as the first step of a fully automated
morphological learning program, to be followed, for example, by a rule
induction procedure that extracts correspondence patterns from paired
forms. See the last section of this paper for further discussion of
possible applications.

We tested our model with German and English input. Our results 
indicate that the algorithm is able to identify a number of pairs 
related by a variety of derivational and inflectional processes with 
a remarkably high precision rate. The algorithm is also 
discovering morphological relationships (such as German plural 
formation with umlaut) that would probably be harder to discover using 
affix-based approaches.

The remainder of the paper is organized as follows: In section
\ref{sec:related}, we shortly review related work.  In
section \ref{sec:model}, we present our model.  In section \ref{sec:evaluation}, we discuss the
results of experiments with German and English input.  Finally, in
section \ref{sec:conclusion} we summarize our main results, we sketch
possible directions that our current work could take, and we discuss
some potential uses for the output of our algorithm.

\section{Related work}\label{sec:related}

For space reason, we discuss here only three approaches that are 
closely related to ours. See, for example, \newcite{Goldsmith:2001} 
for a very different (possibly complementary) approach, and for a 
review of other relevant work.

\subsection{\newcite{Jacquemin:1997}}\label{subsec:jacquemin}

Jacquemin \shortcite{Jacquemin:1997} presents a model that
automatically extracts morphologically related forms from a list of English two-word
medical terms and a corpus from the medical domain.

The algorithm looks for correspondences
between two-word terms and orthographically similar
pairs of words that are adjacent in the corpus.  For example, the
list contains the term \emph{artificial ventilation}, and the corpus
contains the phrase \emph{artificially ventilated}. Jacquemin's algorithm thus 
postulates the (paired) morphological analyses \emph{artificial 
ventilat-ion} and  \emph{artificial-ly ventilat-ed}.

Similar words, for the purposes of this pairing procedure, are simply 
words that share a common left substring (with constraints that we do
not discuss here).

Jacquemin's procedure then builds upon these early steps by clustering
together sets that follow the same patterns, and
using these larger classes to look for spurious analyses.  Finally, the
algorithm tries to cluster classes that are related by similar, rather
than identical, suffixation patterns. Again, we will not describe here
how this is accomplished.

Our basic idea is related to that of Jacquemin, but we propose an 
approach that is more general both in terms of orthography and in 
terms of semantics. In terms of orthography, we do not require that 
two strings share the left (or right) substring in order to constitute 
a candidate pair. Thus, we are not limited to affixal morphological 
patterns. Moreover, our algorithm extracts semantic information 
directly from the input corpus, and thus it does not require a 
pre-compiled list of semantically related pairs. 

\subsection{\newcite{Schone:Jurafsky:2000}}\label{subsec:schonejurafksy}

Schone and Jurafsky \shortcite{Schone:Jurafsky:2000} present a knowledge-free unsupervised
model in which orthography-based distributional cues are combined with
semantic information automatically extracted from word co-occurrence
patterns in the input corpus.

They first look for potential suffixes by searching for frequent word-final substrings.  Then, they look for
potentially morphologically related pairs, i.e., pairs that end in
potential suffixes and share the left substring preceding those
suffixes.  Finally, they look, among those pairs, for those whose
semantic vectors (computed using latent semantic analysis) are
significantly correlated. In short, the idea behind the semantic 
component of their model is that words that tend to co-occur with the 
same set of words, within a certain window of text, are likely to be 
semantically correlated words.

While we follow Schone and Jurafsky's idea of combining orthographic 
and semantic cues, our algorithm differs from them in both respects. 
From the point of view of orthography, we rely on the comparison 
between individual word pairs, without requiring that the two pairs
share a frequent affix, and indeed without requiring that
they share an affix at all.

From the point of view of semantics, we compute scores based on mutual
information instead of latent semantic analysis. Thus, we only
look at the co-occurrence patterns of target words, rather than at
the similarity of their contexts.

Future research should try to assess to what extent these two approaches
produce significantly different results, and/or to what extent they
are complementary.

\subsection{\newcite{Yarowksy:Wicentowski:2000}}\label{subsec:yarowskywicentowski}

Yarowsky and Wicentowski \shortcite{Yarowksy:Wicentowski:2000} propose
an algorithm that extracts morphological rules relating roots and
inflected forms of verbs (but the algorithm can be extended to other
morphological relations).

Their algorithm performs unsupervised, but not completely knowledge-free, learning.  It requires a table of
canonical suffixes for the relevant parts of speech of the target language, a
list of the content word roots with their POS (and some information
about the possible POS/inflectional features of other words), a list
of the consonants and vowels of the language, information about some
characteristic syntactic patterns and, if available, a list of function words.

The algorithm uses a combination of different probabilistic models to find pairs
that are likely to be morphologically related.  One model matches
root + inflected form pairs that have a similar frequency profile. 
Another model matches root + inflected form pairs that tend to co-occur with 
the same subjects and objects (identified using simple regular 
expressions).  Yet another model looks for words that are orthographically similar, in terms of a minimum edit distance score that penalizes consonant changes more than vowel 
changes. Finally, the rules relating stems and inflected forms that
the algorithm extracts from  the pairs it finds in an iteration are
used as a fourth probabilistic model in the subsequent iterations.

Yarowsky and Wicentowski show that the algorithm is extremely accurate 
in identifying English root + past tense form pairs, including those 
pairs that are related by non-affixal patterns (e.g., \emph{think/thought}.)

The main issue with this model is, of course, that it cannot be applied to a new target language without having some \emph{a priori} knowledge about some of its linguistic properties. Thus, the algorithm cannot be applied in cases in which the grammar of the target language has not been properly described yet, or 
when the relevant information is not available for other
reasons. Moreover, even when such information is in principle
available, trying to determine to what extent morphology could be
learned without relying on any other knowledge source remains an
interesting theoretical pursuit, and one whose answer could shed some
light on the problem of human language acquisition.

\section{The current approach: Morphological relatedness as a function
  of orthographic and semantic similarity}\label{sec:model}

The basic intuition behind the model presented here is extremely
simple: Morphologically related words tend to be both orthographically
and semantically similar.  Obviously, there are many
words that are orthographically similar, but are not morphologically
related; for example, \emph{blue} and \emph{glue}. At the same time, 
many semantically related words are not morphologically related (for 
example, \emph{blue} and \emph{green}). However, if two words have a 
similar shape and a related meaning (e.g., \emph{green} and 
\emph{greenish}), they are very likely to be also morphologically related.

In order to make this idea concrete, we use minimum edit distance to
identify words that are orthographically similar, and mutual
information between words to identify semantically related words.

\subsection{Outline of the procedure}\label{subsec:procedure}

Given an unannotated input corpus, the algorithm (after some
elementary tokenization) extracts a list of candidate content words. 
This is simply a list of all the alphabetic
space- or punctuation-delimited strings in the corpus that have a
corpus frequency below .01\% of the total token
count.\footnote{In future versions of the algorithm, we plan to make
  this high frequency threshold dependent on the size of the input corpus.}

Preliminary experiments indicated that our procedure does not perform
as well without this trimming. Notice in
any case that function words tend to be of little morphological
interest, as they display highly lexicalized, often suppletive
morphological patterns.

The word list extracted as described above and the input corpus are
used to compute two lists of word pairs: An \emph{orthographic
similarity list}, in which the pairs are scored on the basis of their
minimum edit distance, and a \emph{semantic similarity list}, based on
mutual information.  Because of minimum thresholds that are
enforced during the computation of the two measures, neither list
contains all the pairs that can in principle be constructed from the input
list.

Before computing the combined score, we get rid of the pairs that do not 
occur in both lists (the rationale being that we do not want to guess the 
morphological status of a pair on the sole basis of orthographic or 
semantic evidence).

We then compute a weighted sum of the orthographic and semantic similarity 
scores of each remaining pair. In the experiments reported below, the 
weights are chosen so that the maximum weighted scores for the two 
measures are in the same order of magnitude (we prefer to align maxima rather 
than means because both lists are trimmed at the bottom, making means 
and other measures of central tendency less meaningful).

The pairs are finally ranked on the basis of the resulting combined 
scores.

In the next subsections, we describe how the orthographic and 
semantic similarity lists are constructed, and some properties of 
the measures we adopted.

\subsection{Scoring the orthographic similarity of word pairs}\label{subsec:orthosim}

Like Yarowsky and Wicentowski, we use minimum edit distance to measure
orthographic similarity.  The minimum edit distance between two
strings is the minimum number of editing operations (insertion, deletion,
substitution) needed to transform one string into the other (see
section 5.6 of \newcite{Jurafsky:Martin:2000} and the references 
quoted there).

Unlike Yarowsky and Wicentowski, we do not attempt to define a 
phonologically sensible edit distance scoring function, as this 
would require making assumptions about how the phonology of the target 
language maps onto its orthography, thus falling outside the domain of 
knowledge-free induction. Instead, we assign a cost of $1$ to all 
editing operations, independently of the nature of the source and
target segments. Thus, in our  system, the pairs \emph{dog/Dog}, \emph{man/men}, \emph{bat/mat} and \emph{day/dry} are all assigned a minimum edit distance of
$1$.\footnote{Following a suggestion by two reviewers, we are
  currently experimenting with an iterative version of our algorithm,
  along the lines of the one described by Yarowsky and Wicentowski. We
  start with the cost matrix described in the text, but we re-estimate
  the editing costs on the basis of the empirical
  character-to-character (or character-to-zero/zero-to-character)
  probabilities observed in the output of the previous run of the
  algorithm. Surprisingly, the revised version of the algorithm leads
  to (moderately) worse results than the single-run version described
  in this paper. Further experimentation with edit cost re-estimation
  is needed, in order to understand which aspects of our iterative
  procedure make it worse than the single-run model, and how it could
  be improved.}

Rather than computing absolute minimum edit distance, we normalize 
this measure by dividing it by the length of the longest string (this 
corresponds to the intuition that, say, two substitutions are less 
significant if we are comparing two eight-letter words than if we are 
comparing two three-letter words). Moreover, since we want to rank 
pairs on the basis of orthographic similarity, rather than 
dissimilarity, we compute (1 - normalized minimum edit 
distance), obtaining a measure that ranges from 1 for identical forms 
to 0 for forms that do not share any character.

This measure is computed for all pairs of words in the potential 
content word list. However, for reasons of size, only pairs that have 
a score of $.5$ or higher (i.e., where the two 
members share at least half of their characters) are recorded in the 
output list.

Notice that orthographic similarity does not favor concatenative affixal
morphology over other types of morphological processes.  For example,
the pairs \emph{woman/women} and \emph{park/parks} both have an
orthographic similarity score of $.8$.

Moreover, orthographic similarity depends only on the two words being
compared, and not on global distributional properties of these words 
and their substrings. Thus, words related by a rare morphological 
pattern can have the same score as words related by a very frequent 
pattern, as long as the minimum edit distance is the same. For 
example, both \emph{nucleus/nuclei} and \emph{bench/benches} have an 
orthographic similarity score of $.714$, despite the fact that the 
latter pair reflects a much more common pluralization pattern.

Of course, this emancipation from edge-anchored concatenation and 
global distributional salience also implies that orthographic similarity will 
assign high scores to many pairs that are \emph{not} morphologically 
related -- for example, the pair \emph{friends/trends} also has an 
orthographic similarity score of $.714$.

Furthermore, since in most languages the range of possible word
lengths is narrow, orthographic similarity as a ranking measure tends
to suffer of a ``massive tying'' problem.  For example, when pairs from
the German corpus described below are ranked on the sole basis of
orthographic similarity, the resulting list is headed by a block of
 19,597 pairs that all have the same score.  These are all pairs where one
word has 9 characters, the other 9 or 8 characters, and the two
differ in only one character.\footnote{Most of the pairs in this
block -- 78\% -- are actually morphologically related.  However,
given that all pairs contain words of length 9 and 8/9 that differ in
one character only, they are bound to reflect only a very small subset of the morphological processes present in German.}

For the above reasons, it is crucial that orthographic similarity is
combined with an independent measure that allows us to distinguish between
similarity due to morphological relatedness vs. similarity due to
chance or other reasons.

\subsection{Scoring the semantic similarity of word 
pairs}\label{subsec:semsim}

Measuring the semantic similarity of words on the basis of raw corpus 
data is obviously a much harder task than measuring the orthographic 
similarity of words.

\emph{Mutual information} (first introduced to computational
linguistics by \newcite{Church:Hanks:1989}) is one of many measures
that seems to be roughly correlated to the degree of semantic
relatedness between words.  The mutual information between two words
$A$ and $B$ is given by:

\begin{equation} \label{mi}
I(A,B) = \log{\frac{Pr(A,B)}{Pr(A)Pr(B)}}
\end{equation}

Intuitively, the larger the deviation between the empirical frequency
of co-occurrence of two words and the expected frequency of
co-occurrence if they were independent, the more likely it is that the
occurrence of one of the two words is \emph{not} independent from the
occurrence of the other.

\newcite{Brown:etal:1990} observed that when mutual information is 
computed in a bi-directional fashion, and by counting co-occurrences of words within a relatively large 
window, but excluding ``close'' co-occurrences (which 
would tend to capture collocations and lexicalized phrases), the 
measure identifies semantically related pairs.

It is particularly interesting for our purposes that most of the
examples of English word clusters constructed on the basis of this
interpretation of mutual information by Brown and colleagues (reported
in their table 6) include morphologically related words.  A similar
pattern emerges among the examples of German words clustered in a
similar manner by \newcite{Baroni:etal:2002}. 
\newcite{Rosenfeld:1996} reports that morphologically related pairs
are common among words with a high (average) mutual information.

We computed mutual information by considering, for each pair, only
co-occurrences within a maximal window of 500 words and outside a minimal
window of 3 words. Given that mutual information is
notoriously unreliable at low frequencies (see, for example,
\newcite{Manning:Schutze:1999}, section 5.4), we only collected mutual
information scores for pairs that co-occurred at least three times (within the relevant window) in the input corpus. Obviously, occurrences across article boundaries were not counted. 
Notice however that the version of the Brown corpus we used does not mark article boundaries. Thus, in this case the whole corpus 
was treated as a single article.

Our ``semantic'' similarity measure is based on the notion that related
words will tend to often occur in the nears of each other.  This
differs from the (more general) approach of \newcite{Schone:Jurafsky:2000}, who look
for words that tend to occur in the same context.  It remains an open
question whether the two approaches produce complementary or
redundant results.\footnote{We are currently experimenting with a measure based on
semantic context similarity (determined on the basis of class-based
left-to-right and right-to-left bigrams), but the current
implementation of this requires \emph{ad hoc} corpus-specific settings
to produce interesting results with both our test corpora.}

Taken by itself, mutual information is a worse predictor of
morphological relatedness than minimum edit distance.  For example,
among the top one hundred pairs ranked by mutual information in each language, only one
German pair and five English pairs are morphologically motivated. This poor performance is not too surprising, given that there are plenty of words that often co-occur together without
being morphologically related. Consider for example (from our English
list) the pairs \emph{index/operand} and \emph{orthodontist/teeth}. 

\section{Empirical evaluation}\label{sec:evaluation}

\subsection{Materials}\label{subsec:materials}

We tested our procedure on the German APA corpus, a corpus of
newswire containing over twenty-eight million word tokens, and on the English Brown
corpus \cite{Kucera:Francis:1967}, a balanced corpus containing 
less than one million two hundred thousand word tokens. Of course, the
most important difference between these two corpora is  that they
represent different languages. However, observe also that they have
very different sizes, and that they are different in terms of the
types of texts constituting them.

Besides the high frequency trimming procedure described above, for 
both languages we removed from the potential content word lists 
those words that were not recognized by the XEROX morphological 
analyzer for the relevant language. The 
reason for this is that, as we describe below, we use this tool to 
build the reference sets for evaluation purposes. Thus, 
morphologically related pairs composed of words not recognized by the 
analyzer would unfairly lower the precision of our algorithm.

Moreover, after some preliminary experimentation, we also decided to
remove words longer than 9 characters from the German list (this
corresponds to trimming words whose length is one standard deviation
or more above the average token length). This actually \emph{lowers} the performance of our system, but makes the
results easier to analyze -- otherwise, the top of the German list
would be cluttered by a high number of rather uninteresting
morphological pairs formed by inflected forms from the paradigm of
very long nominal compounds (such as \emph{Wirtschaftsforschungsinstitut} `institute for economic
research').

Unlike high frequency trimming, the two operations we 
just described are meant to facilitate empirical evaluation, and they do 
not constitute necessary steps of the core algorithm.

\subsection{Precision}\label{subsec:precision}

In order to evaluate the precision obtained by our procedure, 
we constructed a list of all the pairs that,
according to the analysis provided by the XEROX analyzer for the
relevant language, are morphologically related (i.e., share one of
their stems).\footnote{The XEROX morphological analyzers are
  state-of-the-art, knowledge-driven morphological analysis tools (see
  for example \newcite{Karttunen:etal:1997}).} We refer to the lists
constructed in the way we just described as \emph{reference
  sets}.

The XEROX tools we used do not provide derivational analysis for English, and a limited form of derivational
analysis for German.  Our algorithm, however, finds both 
inflectionally and derivationally related pairs. Thus, basing our 
evaluation on a comparison with the XEROX parses leads to an 
underestimation of the precision of the algorithm. We found that this 
problem is particularly evident in English, since English, unlike 
German, has a rather poor inflectional morphology, and thus the 
discrepancies between our output and the analyzer parses in terms of 
derivational morphology have a more visible impact on the results of
the comparison. For example, the English analyzer does not treat pairs related by the adverbial 
suffix \emph{-ly} or by the prefix \emph{un-} as morphologically 
related, whereas our algorithm found pairs such as \emph{soft/softly} 
and \emph{load/unload}.

In order to obtain a more fair assessment of the algorithm, we went 
manually through the first 2,000 English pairs found by our algorithm 
but not parsed as related by the analyzer, looking for items to be
added to the reference set. We were extremely 
conservative, and we added to the reference 
set only those pairs that are related by a transparent and 
synchronically productive morphological pattern. When in doubt, we 
did not correct the analyzer-based analysis. Thus, for example, we did 
\emph{not} count pairs such as \emph{machine/machinery}, 
\emph{variables/varies} or \emph{electric/electronic} as related.

We did not perform any manual post-processing on the German 
reference set.

Tables 1 and 2 report percentage
precision (i.e., the percentage of pairs that 
are in the reference set over the total number of ranked pairs 
up to the relevant threshold) at various cutoff points, for 
German and English respectively.

\begin{table}[htb]
    \small{
    \begin{center}
\begin{tabular}{|r|r|}
    \hline
    \# of pairs &  precision\\
    \hline
    500 & 97\%\\
    1000 & 96\%\\
    1500 & 96\%\\
    2000 & 94\%\\
    3000 & 81\%\\
    4000 & 65\%\\
    5000 & 53\%\\
    5279 & 50\%\\
    \hline
    \end{tabular}
    \label{tab:gerprec}
    \caption{German precision at various cutoff points (5279 = total
      number of pairs)}
    \end{center}
    }
    \end{table}

 \begin{table}[htb]
   \small{
 \begin{center}
     \begin{tabular}{|r|r|}
    \hline
    \# of pairs &  precision\\
    \hline
    500 & 98\%\\
    1000 & 95\%\\
    1500 & 91\%\\
    2000 & 83\%\\
    3000 & 72\%\\
    4000 & 58\%\\
    5000 & 48\%\\
    8902 & 29\%\\
    \hline
    \end{tabular}
    \label{tab:engprec}
    \caption{English precision at various cutoff points (8902 = total
      number of pairs)}
      \end{center}
    }
    \end{table}
%

For both languages we notice a remarkably high precision rate
($>{}90\%)$ up to the 1500-pair cutoff point.

After that, there is a sharper drop in the English precision, whereas
the decline in German is more gradual.  This is perhaps due in part to
the problems with the English reference set we discussed above, but
notice also that English has an overall poorer morphological system and that
the English corpus is considerably smaller than the German one. Indeed, 
our reference set for German contains more than ten times the forms in 
the English reference set.

Notice anyway that, for both languages, the precision rate is still
around 50\% at the 5000-pair cutoff.\footnote{Yarowsky and
  Wicentowski \shortcite{Yarowksy:Wicentowski:2000} report an
    accuracy of over 99\% for their best model and a test set of 3888
    pairs. Our precision rate at a comparable cutoff point is
    much lower (58\% at the 4000-pair cutoff). However, Yarowksy and
    Wicentowski restricted the possible matchings to pairs in which
    one member is an inflected verb form, and the other member is a
    potential verbal root, whereas in our experiments any word in the
    corpus (as long as it was below a certain frequency threshold, and
    it was recognized by the XEROX analyzer) could be matched with any other
    word in the corpus. Thus, on the one hand, Yarowsky and
    Wicentowski forced the algorithm to produce a matching for a
    certain set of words (their set of inflected forms), whereas our
    algorithm was not subject to an analogous constraint. On the other
    hand, though, our algorithm had to explore a much larger possible
    matching space, and it could (and did) make a high number of
    mistakes on pairs (such as, e.g., \textit{sorry} and \textit{worry}) that
    Yarowksy and Wicentowski's algorithm did not have to
    consider. Schone and Jurafsky \shortcite{Schone:Jurafsky:2000}
    report a maximum precision of 92\%. It is hard to compare
    this with our results, since they use a more sophisticated scoring method
    (based on paradigms rather than pairs) and a different type of gold
    standard. Moreover, they do not specify what was the size of the
    input they used for evaluation.}

Of course, what counts as a ``good'' precision rate depends on what we
want to do with the output of our procedure. We show below that even a
very naive morphological rule extraction algorithm can extract
sensible rules by taking whole output lists as its input, since, 
although the number of false positives is high, they are mostly 
related by patterns that are not attested as frequently in the list as 
the patterns relating true morphological pairs. In other words, true
morphological pairs tend to be related by patterns that are
distributionally more robust than those displayed by false
positives. Thus, rule extractors and other procedures processing the
output of our algorithm can probably tolerate a high false positive
rate if they take frequency and other distributional properties of
patterns into account.

Notice that we discussed only precision, and not recall.  This is
because we believe that the goal of a morphological discovery
procedure is not to find the exhaustive list of all morphologically
related forms in a language (indeed, because of morphological
productivity, such list is infinite), but rather to discover all the 
possible (synchronically active and/or common) morphological 
processes present in a language. It is much harder to measure how 
good our algorithm performed in this respect, but the qualitative 
analysis we present in the next subsection indicates that, at least, 
the algorithm discovers a varied and interesting set of morphological
processes.

\subsection{Morphological patterns discovered by the
  algorithm}\label{subsec:rules}

The precision tables confirm that the algorithm found a good number of 
morphologically related pairs. However, if it turned out that all of these 
pairs were examples of the same morphological pattern (say, nominal 
plural formation in \emph{-s}), the algorithm would not be of much
use. Moreover, we stated at the beginning that, since our algorithm
does not assume an edge-based stem+affix concatenation model of
morphology, it should be well suited to discover relations that cannot
be characterized in these terms (e.g., pairs related by
circumfixation, stem changes, etc.). It is interesting to check whether the
algorithm was indeed able to find relations of this sort.

Thus, we performed a qualitative analysis of the output of the
algorithm, trying to understand what kind of morphological processes
were captured by it.

In order to look for morphological processes in the algorithm output, we wrote a program that extracts ``correspondence rules'' in the
following simple way: For each pair, the program looks for the longest
shared (case-insensitive) left- and right-edge substrings (i.e., for a
\emph{stem + suffix} parse and for a \emph{prefix + stem} parse).  The
program then chooses the parse with the longest stem (assuming that
one of the two parses has a non-zero stem), and extracts the relevant
edge-bound correspondence rule.  If there is a tie, the \emph{stem + suffix}
parse is preferred.  The program then ranks the correspondence rules
on the basis of their frequency of occurrence in the original output
list.\footnote{Ranking by cumulative score yields analogous results.}

We want to stress that we are adopting this procedure
as a method to explore the results, and we are by no means proposing 
it as a serious rule induction algorithm. One of the most obvious
drawbacks of the current rule extraction procedure is that it is only
able to extract linear, concatenative, edge-bound suffixation and prefixation
patterns, and thus it misses or fails to correctly generalize some of
the most interesting patterns in the output. Indeed, looking at the 
patterns missed by the algorithm (as we do in part below) is as 
instructive as looking at the rules it found.

Tables 3 and 4 report the top five suffixation and prefixation
patterns found by the rule extractor by taking the entire German and
English output lists as its input.

\begin{table}[tb]
    \small{
    \begin{center}
\begin{tabular}{|l|l|r|}
    \hline
    rule &  example  & fq \\
    \hline
$\epsilon\!\leftrightarrow$s & Jelzin$\leftrightarrow$Jelzins & 921 \\
$\epsilon\!\leftrightarrow$n & lautete$\leftrightarrow$lauteten & 670 \\
$\epsilon\!\leftrightarrow$en & digital$\leftrightarrow$digitalen & 225 \\
$\epsilon\!\leftrightarrow$e & rot$\leftrightarrow$rote & 201 \\
$\epsilon\!\leftrightarrow$es & Papst$\leftrightarrow$Papstes & 113 \\
    \hline
$\epsilon\!\leftrightarrow$ge & stiegen$\leftrightarrow$gestiegen & 9\\
$\epsilon\!\leftrightarrow$\"Ol & Embargo$\leftrightarrow$\"Olembargo & 6\\
$\epsilon\!\leftrightarrow$vor & Mittag$\leftrightarrow$Vormittag & 5\\
aus$\leftrightarrow$ein & ausfuhren$\leftrightarrow$einfuhren & 4\\
ers$\leftrightarrow$drit & Erstens$\leftrightarrow$Drittens & 4\\
    \hline
    \end{tabular}
    \label{tab:gersuff}
    \caption{The most common German suffixation and prefixation patterns}
    \end{center}
    }
    \end{table}

\begin{table}[tb]
    \small{
    \begin{center}
\begin{tabular}{|l|l|r|}
    \hline
    rule &  example  & fq \\
    \hline
$\epsilon\!\leftrightarrow$s & allotment$\leftrightarrow$allotments & 860\\
$\epsilon\!\leftrightarrow$ed & accomplish$\leftrightarrow$accomplished & 98\\
ed$\leftrightarrow$ing & established$\leftrightarrow$establishing & 87\\
$\epsilon\!\leftrightarrow$ing & experiment$\leftrightarrow$experimenting & 85\\
$\epsilon\!\leftrightarrow$d & conjugate$\leftrightarrow$conjugated & 58\\
    \hline
$\epsilon\!\leftrightarrow$un & structured$\leftrightarrow$unstructured & 17\\
$\epsilon\!\leftrightarrow$re & organization$\leftrightarrow$reorganization & 12\\
$\epsilon\!\leftrightarrow$in & organic$\leftrightarrow$inorganic & 7\\
$\epsilon\!\leftrightarrow$non & specifically$\leftrightarrow$nonspecifically & 6\\
$\epsilon\!\leftrightarrow$dis & satisfied$\leftrightarrow$dissatisfied & 5\\
\hline
    \end{tabular}
    \label{tab:engsuff}
    \caption{The most common English suffixation and prefixation patterns}
    \end{center}
    }
    \end{table}

These tables show that our morphological pair scoring procedure found
many instances of various common morphological patterns. With the
exception of the German ``prefixation'' rule
\emph{ers$\leftrightarrow$drit} (actually relating the roots
of the ordinals `first' and `second'), and of the compounding pattern
\emph{$\epsilon\!\leftrightarrow$\"Ol} (`Oil'), all the rules in these
lists correspond to realistic affixation patterns. Not surprisingly,
in both languages many of the most frequent rules (such as, e.g.,
\emph{$\epsilon\!\leftrightarrow$s}) are poly-functional,
corresponding to a number of different morphological relations within and across categories.

The results reported in these tables confirm that the algorithm is
capturing common affixation processes, but they are based on patterns that are
so frequent that even a very naive procedure could uncover
them\footnote{For example, as shown by a reviewer, a procedure that
  pairs words that share the same first five letters, and extracts the
  diverging substrings following the common prefix from each pair.}


More interesting observations emerge from further inspection of the
ranked rule files. For example, among the 70 most frequent German
suffixation rules extracted by the procedure, we encounter those in
table 5.\footnote{In order to find the set of rules presented in table
  5 using the naive algorithm described in the previous footnote, we
  would have to consider the 2672 most frequent rules. Most of these
  2672 rules, of course, do not correspond to true morphological
  patterns -- thus, the interesting rules would be buried in
  noise.}

\begin{table}[htbp]
    \small{
    \begin{center}
\begin{tabular}{|l|l|r|}
    \hline
    rule &  example  & fq \\
    \hline
ag$\leftrightarrow$\"{a}ge & Anschlag$\leftrightarrow$Anschl\"{a}ge & 
10\\
ang$\leftrightarrow$\"{a}nge & R\"uckgang$\leftrightarrow$R\"uckg\"{a}nge & 
6\\
all$\leftrightarrow$\"{a}lle & \"Uberfall$\leftrightarrow$\"Uberf\"{a}lle & 
6\\
ug$\leftrightarrow$\"{u}ge & Tiefflug$\leftrightarrow$Tieffl\"{u}ge & 
5\\
and$\leftrightarrow$\"{a}nde & Vorstand$\leftrightarrow$Vorst\"{a}nde & 
5\\
uch$\leftrightarrow$\"{u}che & Einbruch$\leftrightarrow$Einbr\"{u}che & 
3\\
auf$\leftrightarrow$\"{a}ufe & Verkauf$\leftrightarrow$Verk\"{a}ufe & 
3\\
ag$\leftrightarrow$\"{a}gen & Vertrag$\leftrightarrow$Vertr\"{a}gen & 
3\\

\hline
    \end{tabular}
    \label{tab:umlaut}
    \caption{Some German rules involving stem vowel changes found by
      the rule extractor}
    \end{center}
    }
    \end{table}

The patterns in this table show that our algorithm is capturing the
non-concatenative plural formation process involving fronting of the
stem vowel plus addition of a suffix (\emph{-e/-en}). A smarter rule
extractor should be able to generalize from patterns like these to a
smaller number of more general rules capturing the discontinuous
change. Other umlaut-based patterns that do not involve concomitant
suffixation -- such as in \emph{Mutter/M\"utter} -- were also found by
our core algorithm, but they were wrongly parsed as involving prefixes
(e.g., \emph{Mu$\leftrightarrow$M\"u}) by the rule extractor.

Finally, it is very interesting to look at those pairs that are
morphologically related according to the XEROX analyzer, and that were
discovered by our algorithm, but where the rule extractor could not posit a
rule, since they do not share a substring at either edge. These are
listed, for German, in table 6.

\begin{table}[htbp]
    \small{
    \begin{center}
\begin{tabular}{|l|l|}
    \hline
Alter \"alteren & fordern gefordert\\
Arzt \"Arzte & forderten gefordert\\
Arztes \"Arzte & f\"ordern gef\"ordert\\
Fesseln gefesselt & genannt nannte\\
Folter gefoltert & genannten nannte\\
Putsch geputscht & geprallt prallte\\
Spende gespendet & gesetzt setzte\\
Spenden gespendet & gest\"urzt st\"urzte\\
Streik gestreikt & \\
\hline
    \end{tabular}
    \label{tab:nocommonedge}
    \caption{Morphologically related German pairs that do not share an
      edge found by the basic algorithm}
    \end{center}
    }
    \end{table}

We notice in this table, besides three further instances of
non-affixal morphology, a majority of pairs involving circumfixation
of one of the members.

While a more in-depth qualitative analysis of our results should be
conducted, the examples we discussed here confirm that our algorithm is
able to capture a number of different morphological patterns,
including some that do not fit into a strictly concatenative
edge-bound stem+affix model.
 
\section{Conclusion and Future Directions}\label{sec:conclusion}

We presented an algorithm that, by taking a raw corpus as 
its input, produces a ranked list of morphologically related
pairs at its output. The algorithm finds morphologically related pairs by 
looking at the degree of orthographic similarity (measured by minimum 
edit distance) and semantic similarity (measured by mutual 
information) between words from the input corpus.

Experiments with German and English inputs gave encouraging
results, both in terms of precision, and in terms of the nature of the
morphological patterns found within the output set.

In work in progress, we are exploring various possible improvements 
to our basic algorithm, including iterative re-estimation of edit costs, addition of a context-similarity-based measure, and extension of the output set by 
\emph{morphological transitivity}, i.e. the idea that if word $a$ is 
related to word $b$, and word $b$ is related to word $c$, then word 
$a$ and word $c$ should also form a morphological pair.

Moreover, we plan to explore ways to relax the requirement that all pairs
must have a certain degree of semantic similarity to be treated as
morphologically related (there is evidence that humans treat certain
kinds of semantically opaque forms as morphologically complex -- see
\newcite{Baroni:2000} and the references quoted there). This will
probably involve taking distributional properties of word substrings
into account.

From the point of view of the evaluation of the algorithm, we should
design an assessment scheme that would make our experimental results
more directly comparable to those of
\newcite{Yarowksy:Wicentowski:2000}, \newcite{Schone:Jurafsky:2000}
and others. Moreover, a more in depth qualitative analysis of the
results should concentrate on identifying specific classes of
morphological processes that our algorithm can or cannot identify correctly.

We envisage a number of possible uses for the ranked list that 
constitutes the output of our model. First, the model could provide
the input for a more sophisticated rule extractor, along the lines of
those proposed by \newcite{Albright:Hayes:1999} and
\newcite{Neuvel:2002}. Such models extract morphological
generalizations in terms of correspondence patterns between whole
words, rather than in terms of affixation rules, and are thus
well suited to identify patterns involving non-concatenative morphology
and/or morphophonological changes. A list of related words constitutes
a more suitable input for them than a list of words segmented into morphemes.

Rules extracted in this way would have a number of practical uses --
for example, they could be used to construct stemmers for information
retrieval applications, or they could be integrated into morphological
analyzers.

Our procedure could also be used to replace the first step of
algorithms, such as those of \newcite{Goldsmith:2001} and
\newcite{Snover:Brent:2001}, where heuristic methods are employed to
generate morphological hypotheses, and then an
information-theoretically/probabilistically motivated measure is used
to evaluate or improve such hypotheses. More in
general, our algorithm can help reduce the size of the search space
that all morphological discovery procedures must explore.

Last but not least, the ranked output of (an improved version of) our
algorithm can be of use to the linguist analyzing the morphology of a
language, who can treat it as a way to pre-process her/his data, while still
relying on her/his analytical skills to extract the relevant
morphological generalizations from the ranked pairs.

\subsubsection*{Acknowledgements}

We would like to thank Adam Albright, Bruce Hayes and the anonymous
reviewers for helpful comments, and the Austria Presse Agentur for
kindly making the APA corpus available to us. This work was supported
by the European Union in the framework of the IST programme, project
FASTY (IST-2000-25420). Financial support for \"OFAI is provided by
the Austrian Federal Ministry of Education, Science and Culture.

\end{document}